\documentclass{article}

\usepackage[final]{neurips_2022}

\usepackage[utf8]{inputenc} 
\usepackage[T1]{fontenc}    
\usepackage[backref=section]{hyperref}       
\usepackage{url}              
\usepackage{booktabs}         
\usepackage{amsfonts}         
\usepackage{nicefrac}         
\usepackage{microtype}        
\usepackage[svgnames]{xcolor} 
\usepackage{xkcdcolors}       
\usepackage{amsmath, amsfonts, amssymb}
\usepackage{listings}
\usepackage{float}
\usepackage{graphicx}  
\usepackage{bm}  
\usepackage{wrapfig}
\usepackage{fancyhdr}
\newfloat{lstfloat}{htbp}{lop}
\floatname{lstfloat}{Listing}

\lstdefinelanguage{PythonCustom}{
    keywords=[1]{and, as, assert, break, class, continue, def, del, elif, else, except, exec, finally, for, from, global, if, import, in, is, lambda, not, or, pass, print, raise, return, try, while, with, yield},
    keywordstyle=\color{blue}\bfseries,
    ndkeywords=[2]{compile}, 
    ndkeywordstyle=\color{black}\bfseries,
    sensitive=true,
    morecomment=[l]{\#},
    morestring=[b]',
    morestring=[b]",
}

\renewcommand*\backref[1]{\ifx#1\relax \else (Cited in Section #1) \fi}
\newcommand\norm[1]{\left\lVert#1\right\rVert}

\hypersetup{
  colorlinks=true,
  citecolor=xkcdRoyalBlue,
  linkcolor=xkcdRoyalBlue,
  urlcolor=xkcdRoyalBlue
}

\usepackage{caption}
\title{Nerva: a Truly Sparse Implementation of \\ Neural Networks}

\author{%
  Wieger Wesselink$^1$, Bram Grooten$^1$, Qiao Xiao$^1$, Cassio de Campos$^1$, Mykola Pechenizkiy$^1$\\
  $^1$Eindhoven University of Technology\\
  Eindhoven, The Netherlands \\
  \texttt{\{j.w.wesselink, b.j.grooten, q.xiao, c.decampos, m.pechenizkiy\}@tue.nl} \\
}

\begin{document}

\maketitle


\begin{abstract}
We introduce Nerva, a fast neural network library under development in C++. It supports sparsity by using the sparse matrix operations of Intel's Math Kernel Library (MKL), which eliminates the need for binary masks. 
We show that Nerva significantly decreases training time and memory usage while reaching equivalent accuracy to PyTorch. We run static sparse experiments with an MLP on \mbox{CIFAR-10}. 
On high sparsity levels like $99\%$, the runtime is reduced by a factor of $\bm{4\times}$ compared to a PyTorch model using masks.
Similar to other popular frameworks such as PyTorch and Keras, Nerva offers a Python interface for users to work with.
\end{abstract}

\section{Introduction}

Deep learning models have shown impressive results across several fields of science \citep{brown2020language, fawzi2022discovering, jumper2021highly}. However, these neural networks often come with the drawback of having a very large number of parameters, requiring extensive compute power to train or even test them. To overcome this, researchers have used compression methods to reduce the model size while maintaining performance \citep{han2015learning, cheng2017survey}.

One such compression technique is pruning, where a portion of the weights are removed at the end of the training based on some pre-determined criterion \citep{lecun1989optimal, hassibi1993optimal, han2015learning}. This has led to research into methods for identifying and training sparse networks from the start \citep{frankle2019lottery, lee2019snip, wang2020picking, zhou2019deconstructing, ramanujan2020s, sreenivasan2022rare}.
Further, sparse training methods that adjust the network's topology during training have proven to work well \citep{mocanu2018scalable, bellec2018deep, dettmers2019sparse, evci2020rigging}.

Most of this algorithmic research work is performed with binary masks on top of the weight matrices. The masks enforce sparsity, but the zeroed weights are often still saved in memory and passed in computations. To take full advantage of the sparse algorithms, the sparse neural networks (SNN) community requires truly sparse implementations that show a genuine reduction in compute and memory used.

To solve this issue, we introduce Nerva: a fast neural network library which uses sparse matrix operations, see
\url{https://github.com/wiegerw/nerva}. It is written in C++, but also has a straightforward Python interface for researchers to work with. 
We empirically show that the runtime of Nerva decreases linearly with the model's sparsity level. 
This is an advantage over the default method used by many researchers (i.e., binary masks), which roughly has a constant running time for any sparsity level.

\section{Related Work}

\subsection{Sparse Training}
Sparse training has demonstrated the potential to train efficient networks with sparse connections that match or even outperform their dense counterparts with lower computational costs \citep{mocanu2018scalable, evci2020rigging}.
Starting with \citet{MocanuMNGL16}, it has been shown that initiating a static sparse network without changing its topology during training can also produce comparable performance \citep{lee2019snip, wang2020picking}.
Dynamic Sparse Training, also known as sparse training with dynamic sparsity, is a newer training paradigm that jointly optimizes sparse topology and weights during the training process, starting from a sparse network \citep{mocanu2018scalable, evci2020rigging, yuan2021mest}.
However, most sparse training methods in the literature do not take full advantage of the memory and computational benefits of sparse neural networks and can only achieve theoretical acceleration. This is because they use a binary mask over the connections and depend on dense matrix operations, resulting from the lack of hardware support for sparsity.

\subsection{Truly Sparse Implementations}

To solve the issue of obtaining genuine acceleration in training and inference through sparsity, we need implementations that take advantage of sparse matrix operations.
There are some works that have attempted to implement sparse training in a way that truly saves memory and compute \citep{mocanu2018scalable, curci2021truly, gale2020sparse, elsen2020fast}. For example, the implementation of sparse neural networks with XNNPACK \citep{elsen2020fast} library has shown significant speedups over dense models on smartphone processors. Further, as demonstrated by \citet{liu2021sparse}, sparse training implementations in Cython can effectively conserve memory, enabling the deployment of networks with up to one million neurons on a single laptop.
Another work worth mentioning is DLL \citep{wicht2018dll} which implemented a fast deep learning library in C++. However, DLL does not support sparsity and neither does it have a Python interface, two vital advantages of Nerva.
Lastly, the NVIDIA team is working on hardware that supports sparsity \citep{zhou2021learning, hubara2021accelerated}. In this case usage is quite limited, as it only offers performance benefits for networks with a specific N:M sparsity pattern and is restricted to specific device support.
In Nerva, we aim to improve upon the existing implementations by programming directly in C++, and sidestepping the Python to C conversion.


\section{Background}

In sparse neural networks the goal is to obtain models with as few parameters as possible, while still achieving good performance. The fraction of weights that is removed in comparison to a dense model is given by the global sparsity level $s$, which is the opposite of density $d$
\begin{equation*}
    s = 1 - d 
\end{equation*}
such that a density of $0.01$ corresponds to a sparsity of $0.99$ (or $99\%$).
The density $d^l$ of layer $l$ is given by
\begin{equation*}
    d^l = \frac{ \norm{W^l}_0 }{ n^l_{in} \cdot n^l_{out}  }
\end{equation*}
where $\norm{\cdot}_0$ is the L0-norm, counting the number of non-zero entries in the sparse weight matrix $W^l$. The number of neurons coming in and going out of layer $l$ are given by $n^l_{in}$ and $n^l_{out}$ respectively. The global density $d$ of the model is given by
\begin{equation*}
    d = \frac{\sum_{l=1}^L \norm{W^l}_0 }{\sum_{l=1}^L n^l_{in} n^l_{out}} 
\end{equation*}
where $L$ is the total number of layers. Note that we do not sparsify the biases of each layer, as is often done in the literature.

\section{Implementation}
The Nerva library, written in C++, is a neural network library that aims to provide native support for sparse neural networks.
It includes features such as multilayer perceptions (MLPs), sparse and dense layers, batch normalization, stochastic gradient descent, momentum, dropout, and commonly used activation and loss functions. The development of Nerva is a work in progress, more features will be added in the future.

Important criteria for the design of Nerva are the following:
\begin{itemize}
    \item Runtime efficiency: the implementation is done in C++.
    \item Memory efficiency: the memory footprint is minimized by using truly sparse layers (i.e. we do not use masking).
    \item Energy efficiency: the implementation is optimized for CPU, although we plan to support GPU as well.
    \item Accessibility: a Python interface is provided just as in other frameworks like PyTorch and Keras.
    \item Open design: Nerva is open source, and the implementation is accompanied by precise specifications in pseudocode.
\end{itemize}
The Eigen\footnote{See \url{https://eigen.tuxfamily.org}} library is used for dense matrices, as it offers efficient code for complex matrix expressions. Additionally, the Intel Math Kernel Library (MKL)\footnote{See \url{https://www.intel.com/content/www/us/en/developer/tools/oneapi/onemkl.html}} is utilized to improve computation speed on the CPU through parallelism and processor capabilities such as vectorization. Although Eigen has a sparse matrix type, the performance was not sufficient in our experiments, so the compressed sparse row (CSR) matrix type of the MKL library is used instead. Python bindings are implemented using Pybind11\footnote{See \url{https://github.com/pybind/pybind11}}.
The following operations on sparse matrices are essential for a fast performance:
\begin{align*}
  A &= S B   &\hspace{-2cm} \text{feedforward} \\  
  A &= S^\top B &\hspace{-2cm} \text{backprop} \\  
  S &= A B^\top  &\hspace{-2cm} \text{backprop} \\  
  S &= \alpha S + \beta T, &\hspace{-2cm} \text{momentum}  %
\end{align*}
where $A$ and $B$ are dense matrices, $S$ and $T$ are sparse matrices, and $\alpha$ and $\beta$ are real numbers. Dense matrices are typically: batches of input and output (or gradients thereof), while sparse matrices often represent the weights or their gradients.

Efficient implementations for the first two operations exist in MKL. The third operation is unique in that we only need to compute the values for the non-zero entries of the left-hand side.
A few strategies are implemented to avoid storing the result of the dense product on the right-hand side entirely in memory. Interestingly, the last operation is not efficiently supported in MKL for the case where S and T have the same non-zero entries. We have made an alternative implementation that operates directly on raw data.

In Listing~\ref{listing:nerva} an example of the Nerva Python interface is given, which should look familiar to users of Keras. More code is shown in Listing~\ref{listing:sgd} of Appendix~\ref{app:code}, which contains a possible implementation of stochastic gradient descent (SGD).

\begin{lstfloat}
\begin{lstlisting}
dataset = load_cifar10()
loss = SoftmaxCrossEntropyLoss()
learning_rate_scheduler = ConstantScheduler(0.01)
manual_seed(1234567)
density = 0.05

model = Sequential()
model.add(BatchNormalization())
model.add(Sparse(1000, density, ReLU(), GradientDescent(), Xavier()))
model.add(Dense(128, ReLU(), Momentum(0.9), Xavier()))
model.add(Dense(64, ReLU(), GradientDescent(), Xavier()))
model.add(Dropout(0.3))
model.add(Dense(10, NoActivation(), GradientDescent(), Xavier()))

model.compile(input_size=3072, batch_size=100)
stochastic_gradient_descent(model, dataset, loss, learning_rate_scheduler,
                            epochs=10, batch_size=100, shuffle=True)
\end{lstlisting}
\caption{An example of training a model using the Nerva Python interface. See Listing~\ref{listing:sgd} in Appendix~\ref{app:code} for an implementation of the \texttt{stochastic\_gradient\_descent} function.}
\label{listing:nerva}
\end{lstfloat}

\section{Experiments}

In this section, we present our experiments comparing Nerva to the popular deep learning framework PyTorch. First we go into our experimental setup, after which we present and interpret the results. Additional graphs are shown in Appendix~\ref{app:graphs}.

\begin{wraptable}{R}{0.305\textwidth}
    \centering
    \caption{Initial learning rates in our experiments.}
    \label{tab:lr}
    \begin{tabular}{rl}
    \toprule
       Density  &  Learning rate\\
    \midrule
       $1$      &  $0.01$   \\
       $0.5$    &  $0.01$   \\
       $0.2$    &  $0.01$   \\
       $0.1$    &  $0.03$   \\
       $0.05$   &  $0.03$   \\
       $0.01$   &  $0.1$    \\
       $0.005$  &  $0.1$    \\
       $0.001$  &  $0.1$    \\
    \bottomrule
    \end{tabular}
\end{wraptable}

\newpage
\subsection{Experimental setup}

We train on the CIFAR-10 dataset \citep{krizhevsky2009learning}, using a standard multilayer perception (MLP) model with layer sizes 
$[3072, 1024, 512, 10]$ and ReLU activations. The weights are initialized with Xavier \citep{glorot2010understanding}. We augment the data in a standard manner often used in the literature.
We use a batch size of $100$ and the SGD optimizer with momentum$=0.9$, Nesterov$=\textsf{True}$, and no weight decay.  
The learning rate starts at a value which depends on the sparsity level (see Table~\ref{tab:lr}) and is decayed twice during training: after 50\% and 75\% of the epochs. We use a decay factor of $0.1$ and train for $100$ epochs.

We run on multiple sparsity levels, from $50\%$ up to $99.9\%$ sparsity, and also compare the performance of the fully dense model. The exact global densities used are shown in Table~\ref{tab:lr}.
We distribute the sparsity levels over the layers according to the 
 Erd\H{o}s-R\'{e}nyi (ER) initialization scheme from \citet{mocanu2018scalable}, which applies higher sparsity levels to larger layers. For instance, for a sparsity level of $99\%$, the density of each layer is as follows: $[0.008, 0.018, 0.6]$. 
 The last layer, which is the smallest, receives the lowest sparsity of $1-0.6=40\%$. 
When the network used in our experiments is fully dense, it has $3,676,682$ parameters. At a sparsity level of $99.9\%$ this drops down to $5,221$ parameters.


We compare our new Nerva framework with PyTorch.
Nerva uses sparse matrix operations, while for PyTorch we apply binary masks, a technique often employed in the sparsity literature. 
We aim for a completely fair comparison between the frameworks. Thus, we attempt to ensure that all the implementation details have exactly the same settings in both frameworks. All experiments are run on the same desktop, see Appendix~\ref{app:hardware} for its specifications. We run 3 random seeds for each choice of framework and density level.

\begin{figure}[H]
    \centering
    \includegraphics[width=0.7\textwidth]{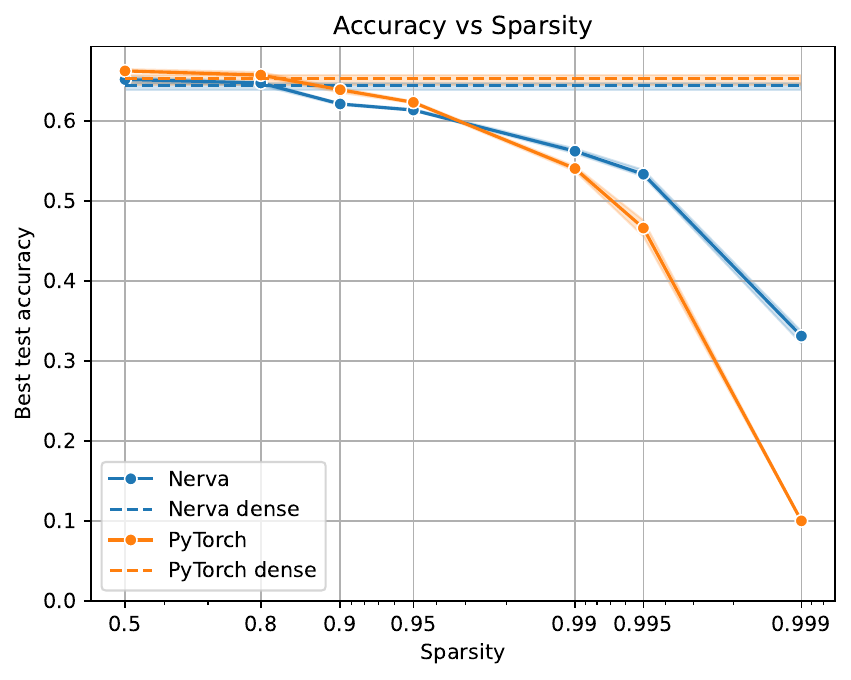}
    \caption{Accuracy vs sparsity. Notice the logit-scale on the horizontal axis, values closer to 1 are stretched out. The accuracy of Nerva and PyTorch are similar, except for the high sparsity regime where Nerva outperforms PyTorch. The reason for this is yet unknown.} 
    \label{fig:acc-vs-sparsity}
\end{figure}

\subsection{Equivalent accuracy}
\label{sec:acc}

We measure the training and test accuracy over time. In Figure~\ref{fig:acc-vs-sparsity} we report the best test accuracy over the entire training run, and plot it against the various global sparsity levels that we used. We used 3 random seeds for each setting, and show the averages with a 95\% confidence interval. The horizontal axis of Figure~\ref{fig:acc-vs-sparsity} has a logit-scale\footnote{The logit function is $\text{logit}(s) = \log(s / (1-s))$. See the \href{https://matplotlib.org/3.1.1/api/scale_api.html\#matplotlib.scale.LogitScale}{matplotlib documentation} for details.} to improve the visibility of high sparsity levels.

The accuracy of Nerva and PyTorch is very similar, which is what we aimed for.
The only exception is the higher sparsity levels, where Nerva seems to outperform PyTorch. We are unsure if this is due to an advantage of truly sparse training, or whether it comes from a tiny discrepancy in implementation details we might have missed.

\subsection{Decreased training time}

\begin{figure}
    \centering
    \includegraphics[width=0.7\textwidth]{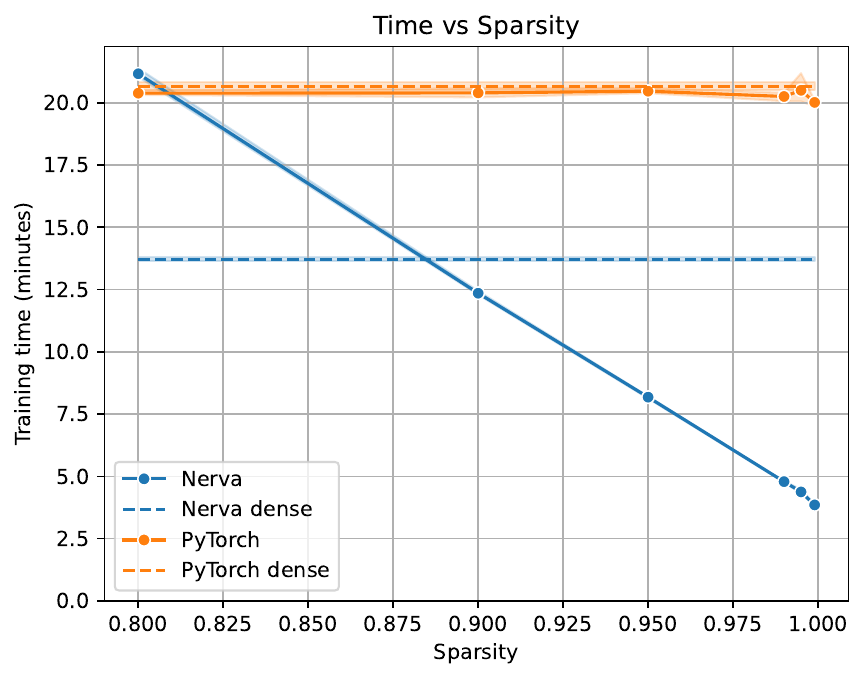}
    \caption{The total training time of 100 epochs for CIFAR-10, on a regular desktop with 4 CPU cores. As the sparsity level increases, the running time of Nerva goes down linearly, as it takes advantage of sparse matrix operations. The running time for PyTorch stays roughly constant, because it uses binary masks.}
    \label{fig:time-vs-sparsity}
\end{figure}

For each epoch we measure how much time it took to perform all the necessary (sparse) operations. We exclude the time needed for loading and augmenting the data. We sum the times of all 100 epochs together, which is what Figure~\ref{fig:time-vs-sparsity} shows. 

As expected, the running time for PyTorch stays approximately constant (independent of the sparsity level) as this implementation uses binary masks. It needs to multiply all weights of each matrix, whether it is sparse or not. On the contrary, Nerva shows its true advantage here. As the sparsity level goes up, the running time decreases linearly. Less multiplications are necessary, and this drop in total FLOPs is reflected in a considerable reduction in running time.

\subsection{Decreased inference time}
\label{sec:inference}

\begin{figure}[t]
    \centering
    \includegraphics[width=0.7\textwidth]{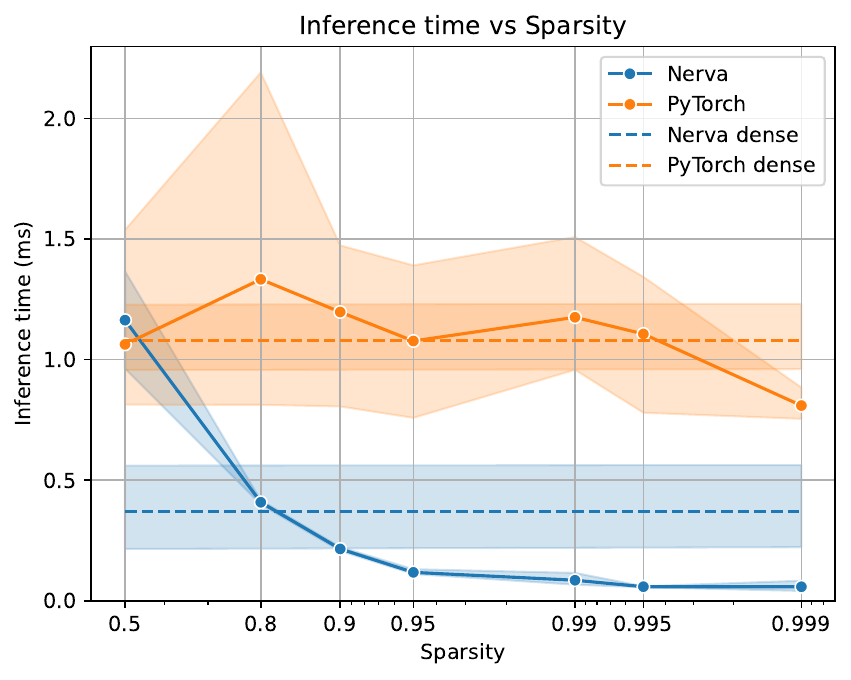}
    \caption{Inference time vs sparsity. The graph shows the average inference time of 1 example of CIFAR-10 in milliseconds, on a regular desktop with 4 CPU cores. 
    Like in figure \ref{fig:acc-vs-sparsity} a logit-scale is used. The inference time of Nerva
    is significantly lower, especially for higher sparsity levels.
    \label{fig:inference-vs-sparsity}}
\end{figure}

We measure the inference time needed for one example of CIFAR-10. The computation is done with batch size 1. In Figure~\ref{fig:inference-vs-sparsity} we plot the inference time against the various global sparsity levels that we used. We used 3 random seeds for each setting, and show the averages with a 95\% confidence interval. As in the previous sections we use a logit-scale to improve the visibility of high sparsity levels.

The inference time that we measured for Nerva is significantly lower than for PyTorch, with the exception of the very low sparsity levels. As expected, for higher sparsity levels the inference time decreases significantly.

\subsection{Scalability}
To measure the scalability of our sparse neural network solution, we did a few experiments that should give an indication of the running time for larger models. Table \ref{table:scalability}, left shows the training times of one epoch for CIFAR-10 using a sparse model with density $0.01$ and a varying number of hidden layers of size $1024$. Overall the Nerva model runs about $4\times$ faster, and the runtime scales linearly in the number of hidden layers. Table \ref{table:scalability}, right shows the runtime in the case of three equally sized hidden layers, with sizes ranging from $1024$ to $32,768$. 
Again in all cases Nerva is faster. However, the factor between Nerva and PyTorch drops from $4\times$ to $1.5\times$ for the large matrices. This is because the dense matrix multiplication routines of the MKL library happen to scale much better for large matrices than their sparse equivalents. Note that for size $32,768$ the PyTorch model ran out of memory (i.e., over 32GB), while the Nerva model was still only using around 2GB.

\begin{table}
    \centering
        \caption{The running times in seconds of 1 epoch for CIFAR-10 using sparse models with density 0.01. On the \textbf{left} we increase the depth ($N$), showing results for $N$ hidden layers of size 1024, where $N$ ranges from 1 to 10. On the \textbf{right} we adjust the width ($M$), comparing results for three hidden layers of size $M$, with $M$ ranging from 1024 to 32768 (where PyTorch runs out of memory).}
\label{table:scalability}
\vspace{0.1cm}
    \begin{tabular}{rlll}
    \toprule
     depth & Nerva & PyTorch & factor \\
    \midrule
      1 &  2.37 &   10.53 &   4.44$\times$ \\
      2 &  3.29 &   13.92 &   4.23$\times$ \\
      3 &  4.20 &   17.58 &   4.19$\times$ \\
      4 &  5.08 &   21.94 &   4.32$\times$ \\
      5 &  5.95 &   24.99 &   4.20$\times$ \\
      6 &  6.89 &   28.90 &   4.20$\times$ \\
      7 &  7.79 &   32.78 &   4.21$\times$ \\
      8 &  8.71 &   36.19 &   4.15$\times$ \\
      9 &  9.56 &   39.75 &   4.16$\times$ \\
     10 & 10.45 &   43.11 &   4.13$\times$ \\
    \bottomrule
    \end{tabular}
    \hspace{0.7cm}
    \begin{tabular}{rrrr}
    \toprule
     width &   Nerva & PyTorch & factor \\
    \midrule
     1024 &    4.20 &   17.58 &   4.19$\times$ \\
     2048 &   10.23 &   48.06 &   4.70$\times$ \\
     4096 &   35.53 &  137.54 &   3.87$\times$ \\
     8192 &  128.87 &  456.12 &   3.54$\times$ \\
    16384 & 1100.56 & 1669.90 &   1.52$\times$ \\
    32768 & 4466.92 &    - &   - \\
    \bottomrule
    \end{tabular}
\end{table}

\subsection{Memory}

To estimate the memory consumption of the Nerva sparse models, we store the weights as NumPy tensors in .npy format. For dense layers, we save one tensor containing the weight values. The weight values of sparse layers are stored in CSR format, which means that for each non-zero entry, two additional integers are saved: a row and column index. Hence for sparse layers, we store a vector containing the non-zero values and two vectors containing the column and row indices. We applied this storage scheme to a CIFAR-10 model with hidden layers sizes $1024$ and $512$. The disk sizes for multiple densities are shown in Table~\ref{tab:memory}.
The difference between these sizes should be a rough indicator of the memory requirements of these models. In particular, for a sparse model with density $0.01$ a $49\times$ reduction is achieved compared to the fully dense model.


\begin{table}[H]
    \centering
    \captionsetup{width=.66\textwidth,skip=0.2cm}
    \caption{The memory required by Nerva to save the MLP model (layers: 3072-1024-512-10) used in our experiments.}
    \label{tab:memory}
    \begin{tabular}{rr}
        \toprule
        Density & Memory size \\
        \midrule
        1 & 15MB \\
        0.1 & 2.8MB \\
        0.01 & 295KB \\
        0.001 & 37KB \\
        \bottomrule
    \end{tabular}
\end{table}

\section{Discussion and Conclusion}

We present a new library, Nerva, for fast computations in sparse neural networks (SNN). From the results we see that under certain sparsity levels, i.e., above $\sim 80\%$, Nerva outperforms PyTorch in running time, while achieving equivalent accuracy. 
This particular threshold (sparsity level $0.8$) where Nerva surpasses PyTorch in efficiency is dependent on the size of the model and the datasets trained on. This is a promising case for sparsity in the light of today's scaling laws, because in our preliminary experiments we see that \textit{the larger the model, the lower the sparsity level needed} for Nerva to beat PyTorch in efficiency.

\paragraph{Limitations \& Future Work}
The presented experiments on sparse networks all use a static sparse topology structure. We will add functionality for Dynamic Sparse Training to Nerva and intend to report on its results in the future. Note that our experiments in this work are all run on CPUs. We plan to report further on GPU results in the near future. At this moment it is an open question whether a sparse GPU implementation (based on MKL) is able to compete with a dense GPU implementation.
We will open-source our Nerva implementation on GitHub and encourage everyone to contribute.
We hope to motivate the SNN community to work on increasing the true efficiency of our sparse algorithms.

\newpage
\bibliographystyle{apalike}
\bibliography{references}

\newpage
\appendix
\section*{Appendix}

\section{Hardware specifications}
\label{app:hardware}

We run all our experiments on the same machine for a fair comparison of running times. We use a desktop with 4 CPU cores of the type Intel Core i7-6700 @ 3.40Ghz. The machine has 32 GB of memory and runs a Linux operating system. We have not used GPUs in our experiments, this is reserved for future work.

\section{Code}
\label{app:code}

In Listing~\ref{listing:sgd} we show some more code which complements Listing~\ref{listing:nerva} from the main body.

\begin{lstfloat}[H]
\begin{lstlisting}
def stochastic_gradient_descent(model, dataset, loss, learning_rate, 
                                epochs, batch_size, shuffle):
    N = dataset.Xtrain.shape[1]  # the number of examples
    I = list(range(N))
    K = N // batch_size  # the number of batches
    for epoch in range(epochs):
        if shuffle: random.shuffle(I)
        eta = learning_rate(epoch)  # update the lr at the start of each epoch
        for k in range(K):
            batch = I[k * batch_size: (k + 1) * batch_size]
            X = dataset.Xtrain[:, batch]
            T = dataset.Ttrain[:, batch]
            Y = model.feedforward(X)
            dY = loss.gradient(Y, T) / batch_size
            model.backpropagate(Y, dY)
            model.optimize(eta)
\end{lstlisting}
\caption{An implementation of SGD using the Nerva Python interface.}
\label{listing:sgd}
\end{lstfloat}

\section{Additional plots}
\label{app:graphs}

We provide additional plots to the experiments. 

\newpage
\subsection{Accuracy-vs-epoch}

\begin{figure}[H]
    \centering
    \includegraphics[width=\textwidth]{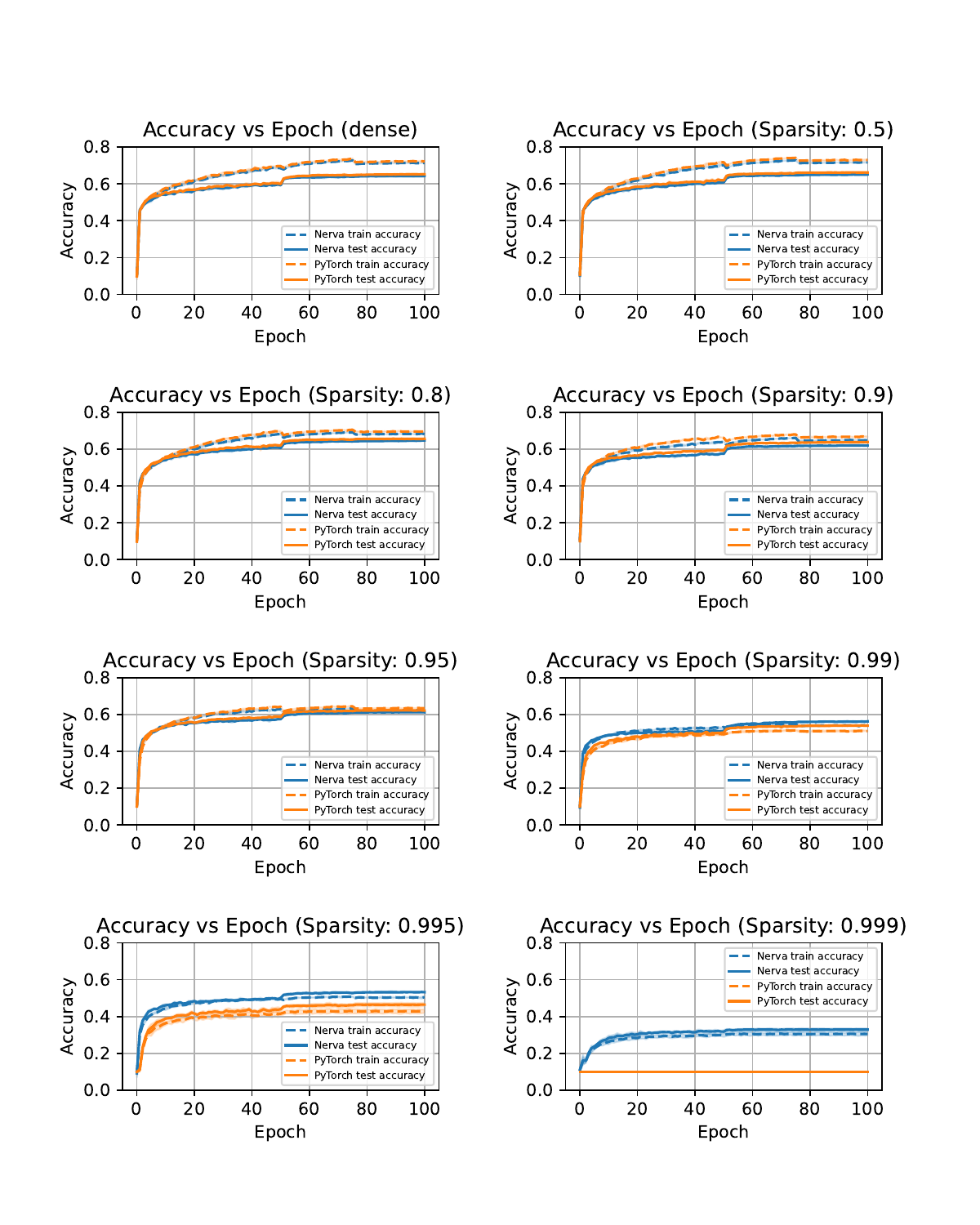}
    \caption{Accuracy vs Epoch. The comparison of the test and training accuracy of Nerva and PyTorch during training on CIFAR-10 with various sparsity levels, over three runs with different seeds.}
    \label{fig:accuracy}
\end{figure}

\subsection{Loss-vs-epoch}

\begin{figure}[H]
    \centering
    \includegraphics[width=\textwidth]{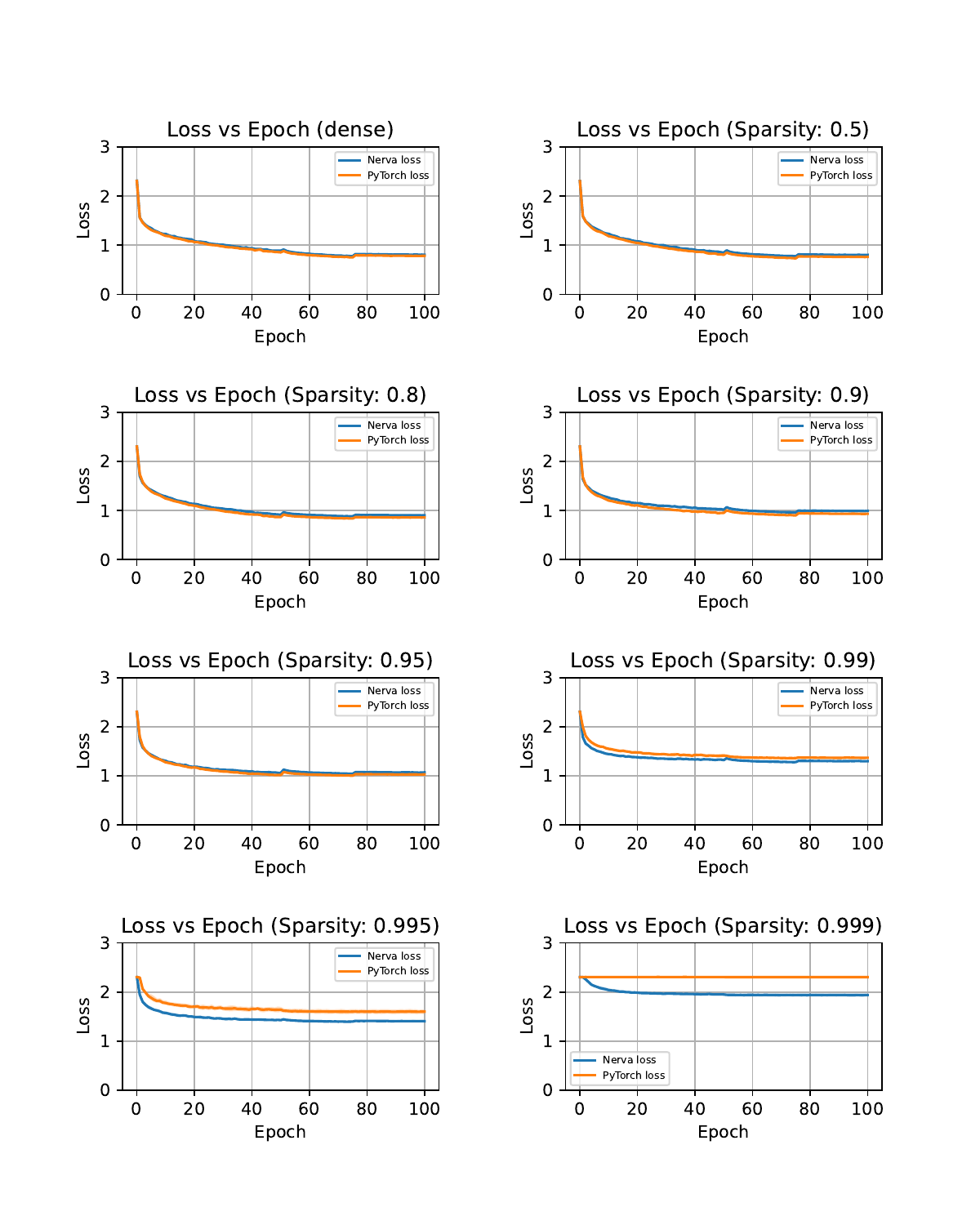}
    \caption{Loss vs Epoch. The comparison of learning curves of Nerva and PyTorch during training on CIFAR-10 with various sparsity levels, over three runs with different seeds.}
    \label{fig:loss}
\end{figure}

\newpage
\subsection{Runtime}

\begin{figure}[H]
    \centering
    \includegraphics[width=0.7\textwidth]{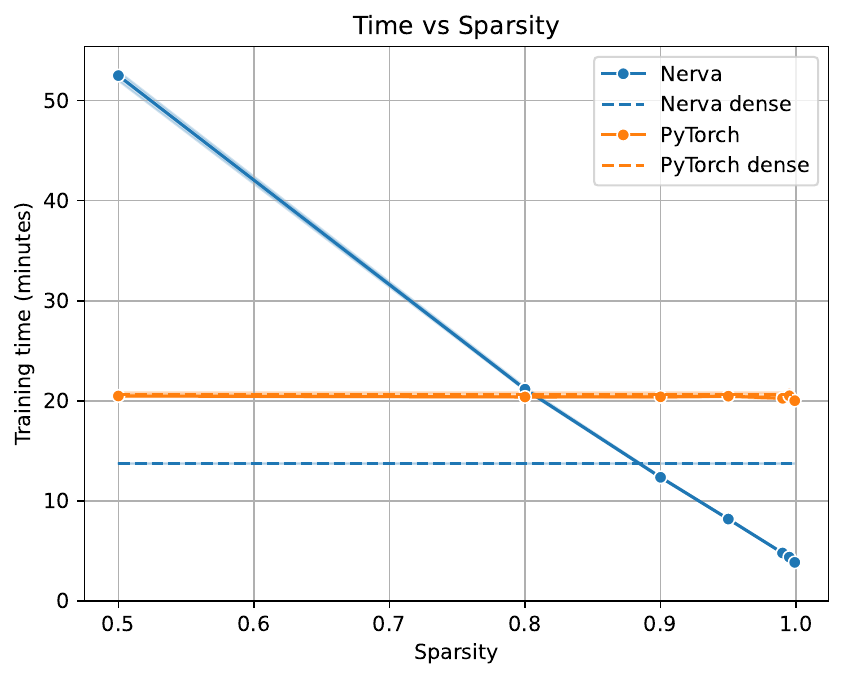}
    \caption{The total training time of 100 epochs for CIFAR-10, on a regular desktop with 4 CPU cores. As the sparsity level increases, the running time of Nerva goes down linearly, as it takes advantage of sparse matrix operations. The running time for PyTorch stays roughly constant, because it uses binary masks.}
    \label{fig:time-vs-sparsity_with50}
\end{figure}

In Figure~\ref{fig:time-vs-sparsity_with50} we show a zoomed-out plot compared to Figure~\ref{fig:time-vs-sparsity} in the main body. It shows that for low sparsity levels like $50\%$ the sparse matrix operations do not show their benefit yet. In these cases using binary masks is faster. We plan to implement both options in Nerva, such that it can always use the fastest option.

\end{document}